\documentclass[11pt,a4paper]{article}
\usepackage[hyperref]{acl2020}
\usepackage{times}
\usepackage{latexsym}

\usepackage{url}
\usepackage{times}
\usepackage{latexsym}
\usepackage{booktabs}
\usepackage{comment}
\usepackage{footnote}
\usepackage{microtype}
\usepackage{amsmath}
\usepackage{graphicx}
\usepackage{cleveref}
\usepackage{tikz}
\usepackage{etoolbox}
\usepackage{enumitem}
\usepackage{xcolor}
\usepackage[ruled,linesnumbered]{algorithm2e}

\newcommand{\rs}[0]{{\bf AIR}} 
\usepackage{microtype}

\aclfinalcopy 

\title{Unsupervised Alignment-based Iterative Evidence Retrieval for Multi-hop Question Answering}

\author{Vikas Yadav, Steven Bethard, Mihai Surdeanu \\
  University of Arizona, Tucson, AZ, USA \\
  {\tt \{vikasy, bethard, msurdeanu\}@email.arizona.edu}}
  
\date{}

\begin{document}
\maketitle
\begin{abstract}
Evidence retrieval is a critical stage of question answering (QA), necessary not only to improve performance, but also to explain the decisions of the corresponding QA method. We introduce a simple, fast, and unsupervised iterative evidence retrieval method, which relies on three ideas:
(a) an unsupervised alignment approach to soft-align questions and answers with justification sentences using only GloVe embeddings, 
(b) an iterative process that reformulates queries focusing on terms that are not covered by existing justifications, which (c) a stopping criterion that terminates retrieval when the terms in the given question and candidate answers are covered by the retrieved justifications.
Despite its simplicity, our approach outperforms all the previous methods (including supervised methods) on the evidence selection task on two datasets: MultiRC and QASC. When these evidence sentences are fed into a RoBERTa answer classification component, we achieve state-of-the-art QA performance on these two datasets.  
\end{abstract}

\section{Introduction}

Explainability in machine learning (ML) remains a critical unsolved challenge that slows the adoption of ML in real-world applications~\cite{biran2017explanation, gilpin2018explaining, alvarez2017causal, arras2017relevant}.

Question answering (QA) is one of the challenging natural language processing (NLP) tasks that benefits from explainability. In particular, multi-hop QA requires the aggregation of multiple evidence facts in order to answer complex natural language questions \cite{yang2018hotpotqa}. 
Several multi-hop QA datasets have been proposed recently \cite{yang2018hotpotqa, khashabi2018looking, welbl2018constructing, dua2019drop, chen2019understanding, khot2019qasc, sun2019dream, jansen2019textgraphs, rajpurkar2018know}. While several neural methods have achieved state-of-the-art results on these datasets \cite{devlin2018bert, liu2019roberta, yang2019xlnet},
we argue that many of these directions lack a human-understandable explanation of their inference process, which is necessary to transition these approaches into real-world applications.
This is especially critical for multi-hop, multiple choice QA (MCQA) where: (a) the answer text may not come from an actual knowledge base passage, and (b) reasoning is required to link the candidate answers to the given question \cite{yadav2019quick}. 
Figure~\ref{QASC_diagram} shows one such multi-hop example from a MCQA dataset.

In this paper we introduce a simple {\bf {\em a}}lignment-based {\bf {\em i}}terative {\bf {\em r}}etriever (\rs)\footnote{\url{https://github.com/vikas95/AIR-retriever}}, which retrieves high-quality evidence sentences from unstructured knowledge bases. We demonstrate that these evidence sentences are useful not only to explain the required reasoning steps that answer a question, but they also considerably improve the performance of the QA system itself.

Unlike several previous works that depend on {\em supervised} methods for the retrieval of justification sentences (deployed mostly in settings that rely on small sets of candidate texts, e.g., HotPotQA, MultiRC), 
\rs\  is completely {\em unsupervised} and scales easily from QA tasks that use small sets of candidate evidence texts to ones that rely on large knowledge bases (e.g., QASC \cite{khot2019qasc}). 
 \rs\ retrieves justification sentences through a simple iterative process. In each iteration, \rs\ uses an alignment model to find justification sentences that are closest in embedding space to the current query~\cite{bridging_the_gap, yadav2018sanity}, which is initialized with the question and candidate answer text. 
 After each iteration, \rs\ adjusts its query to focus on the {\em missing} information \cite{khot2019missing} in the current set of justifications. 
 \rs\ also conditionally expands the query using the justifications retrieved in the previous steps. 

In particular, our key contributions are:
{\flushleft {\bf (1)}}
We develop a simple, fast, and unsupervised iterative evidence retrieval method, which achieves state-of-the-art results on justification selection on two multi-hop QA datasets: MultiRC \cite{khashabi2018looking} and QASC \cite{khot2019qasc}. Notably, our simple unsupervised approach that relies solely on GloVe embeddings \cite{pennington2014glove} outperforms three transformer-based supervised state-of-the-art methods: BERT \cite{devlin2018bert}, XLnet \cite{yang2019xlnet} and RoBERTa \cite{liu2019roberta} on the justification selection task. Further, when the retrieved justifications are fed into a QA component based on RoBERTa \cite{liu2019roberta}, we obtain the best QA performance on the development sets of both MultiRC and QASC.\footnote{In settings where external labeled resources are not used.} 

{\flushleft {\bf (2)}} \rs\ can be trivially extended to capture parallel evidence chains by running multiple instances of \rs\ in parallel starting from different initial evidence sentences. We show that aggregating multiple parallel evidences further improves the QA performance over the vanilla \rs\ by 3.7\% EM0 on the MultiRC and 5.2\% accuracy on QASC datasets (both absolute percentages on development sets). Thus, with 5 parallel evidences from \rs\, we obtain 36.3\% EM0 on MultiRC and 81.0\% accuracy on QASC {\bf hidden} test sets (on their respective leaderboards). To our knowledge from published works, these results are the new state-of-the-art QA results on these two datasets. These scores are also accompanied by new state-of-the-art performance on evidence retrieval on both the datasets, which emphasizes the interpretability of \rs.

{\flushleft {\bf (3)}} 
We demonstrate that \rs's iterative process that focuses on missing information is more robust to semantic drift. 
We show that even the supervised RoBERTa-based retriever trained to retrieve evidences iteratively, suffers substantial drops in performance with retrieval from consecutive hops. 



\begin{figure}[t!]
\small
Question: Exposure to oxygen and water can cause iron to \\

(A) decrease strength (B) melt (C) uncontrollable burning (D) thermal expansion (E) {\bf turn orange on the surface} (F) vibrate (G) extremes of temperature (H)  levitate

\hrulefill

Gold justification sentences:
\begin{enumerate}[nosep]

\item  when a metal rusts , that metal becomes orange on the surface 
 \item Iron rusts in the presence of oxygen and water.
\end{enumerate}
\hrulefill \\
Parallel evidence chain 1:
\begin{enumerate}[nosep]

\item Dissolved oxygen in water usually causes the oxidation of iron.
\item When iron combines with oxygen it turns orange.
\end{enumerate}

Parallel evidence chain 2:
\begin{enumerate}[nosep]
\item By preventing the exposure of the metal surface to oxygen, oxidation is prevented.
\item When iron oxidizes, it rusts.

\end{enumerate}
\caption{\small An example question that requires multi-hop reasoning, together with its gold justifications from the QASC dataset. The two parallel evidence chains retrieved by \rs\ (see \cref{sec:approach}) provide imperfect but relevant explanations for the given question.}
\label{QASC_diagram}
\vspace{-4mm}
\end{figure}

\section{Related Work}

Our work falls under the revitalized direction that focuses on the interpretability of QA systems, where the machine's inference process is explained to the end user in natural language evidence text \cite{qi2019answering,yang2018hotpotqa,wang2019evidence,yadav2019quick,bauer2018commonsense}. Several datasets in support of interpretable QA have been proposed recently. For example, datasets such as HotPotQA, MultiRC, QASC, Worldtree Corpus, etc., \cite{yang2018hotpotqa,khashabi2018looking,khot2019qasc,jansen2019textgraphs} provide annotated evidence sentences enabling the automated evaluation of interpretability via evidence text selection. 

QA approaches that focus on interpretability can be broadly classified into three main categories: {\em supervised}, which require annotated justifications at training time, {\em latent}, which extract justification sentences through latent variable methods driven by answer quality, and, lastly, {\em unsupervised} ones, which use unsupervised algorithms for evidence extraction.

In the first class of supervised approaches, a supervised classifier is normally trained to identify correct justification sentences driven by a query \cite{nie2019revealing,tu2019select,banerjee2019asu}. Many systems tend to utilize a multi-task learning setting to learn both answer extraction and justification selection  with the same network \cite{min2018efficient,gravina2018cross}. Although these approaches have achieved impressive performance, they rely on annotated justification sentences, which may not be always available. 
Few approaches have used distant supervision methods \cite{lin2018denoising,wang2019evidence} to create noisy training data for evidence retrieval but these usually underperform due to noisy labels. 

In the latent approaches for selecting justifications, reinforcement learning \cite{geva2018learning,choi2017coarse} and PageRank \cite{surdeanu2008learning} have been widely used to select justification sentences without explicit training data. While these directions do not require annotated justifications, they tend to need large amounts of question/correct answer pairs to facilitate the identification of latent justifications. 

In unsupervised approaches, many QA systems have relied on structured knowledge base (KB) QA. 
For example, several previous works have used ConceptNet \cite{speer2017conceptnet} to keep the QA process interpretable \cite{khashabi2018question,sydorova2019interpretable}.
However, the construction of such structured knowledge bases is expensive, and may need frequent updates. Instead, in this work we focus on justification selection from textual (or unstructured) KBs, which are inexpensive to build and can be applied in several domains. In the same category of unsupervised approaches, conventional information retrieval (IR) methods such as BM25 \cite{chen2017reading} have also been widely used to retrieve independent individual sentences.
As shown by \cite{khot2019qasc,qi2019answering}, and our \cref{tab:QASC}, these techniques do not work well for complex multi-hop questions, which require knowledge aggregation from multiple related justifications. 
Some unsupervised methods extract groups of justification sentences \cite{chen2019multi,yadav2019quick} but these methods are exponentially expensive in the retrieval step. Contrary to all of these, \rs\ proposes a simpler and more efficient method for chaining justification sentences. 


 
Recently, many supervised iterative justification retrieval approaches for QA have been proposed~\cite{qi2019answering,feldman2019multi,banerjee2019asu,das2018multi}. While these were shown to achieve good evidence selection performance for complex questions when compared to earlier approaches that relied on just the original query \cite{chen2017reading,yang2018hotpotqa}, they all require supervision. 

\begin{figure*} [t!]
\includegraphics[width=16.1cm]{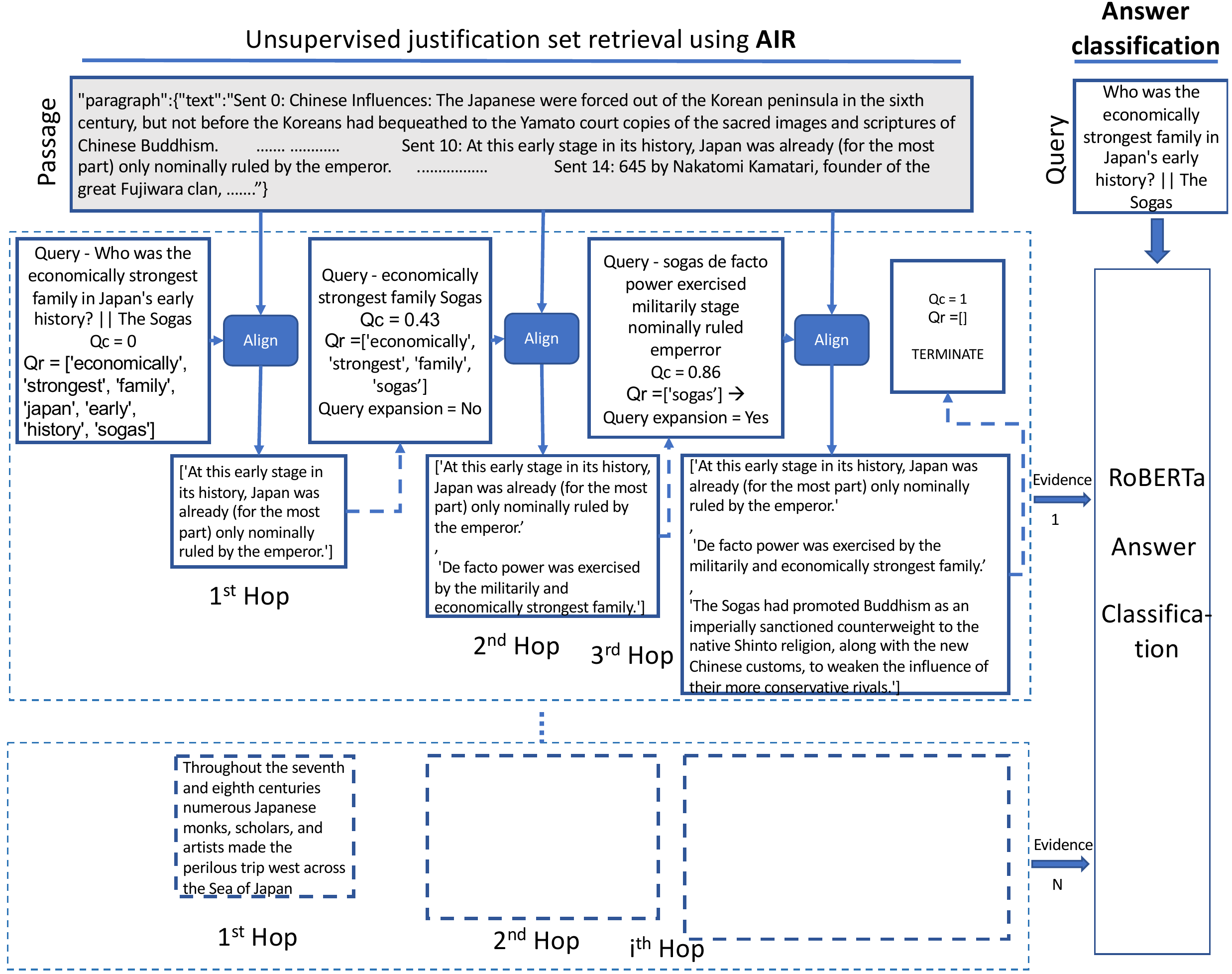}  
\vspace{-\baselineskip}
\caption{\label{fig:AIR_multiRC_example} \small
A walkthrough example showing the iterative retrieval of justification sentences by \rs\ on MultiRC. 
Each current query includes keywords from the original query (which consists of question + candidate answer) that are not covered by previously retrieved justifications (see $2^{nd}$ hop). If the number of uncovered keywords is too small, the query is expanded with keywords from the most recent justification ($3^{rd}$ hop). 
The retrieval process terminates when all query terms are covered by existing justifications. $Q_c$ indicates the proportion of query terms covered in the justifications; $Q_r$ indicates the query terms which are still not covered by the justifications. 
\rs\ can retrieve parallel justification chains by running the retrieval process in parallel, starting from different candidates for the first justification sentence in a chain. 
 }
 \vspace{-4mm}
\end{figure*}

As opposed to all these iterative-retrieval methods and previously discussed directions, our proposed approach \rs\ is completely unsupervised, i.e., it does not require annotated justifications. 
Further, unlike many of the supervised iterative approaches~\cite{feldman2019multi,sun2019pullnet} that perform query reformulation in a continuous representation space, \rs\ employs a simpler and more interpretable query reformulation strategy that relies on explicit terms from the query and the previously retrieved justification.
Lastly, none of the previous iterative retrieval approaches address the problem of semantic drift, whereas \rs\ accounts for drift by controlling the query reformulation as explained in \cref{retrieval_explanation}.

 

\section{Approach}
\label{sec:approach}

As shown in \cref{fig:AIR_multiRC_example}, the proposed QA approach consists of two components: (a) an unsupervised, iterative component that retrieves chains of justification sentences given a query; and (b) an answer classification component that classifies a candidate answer as correct or not, given the original question and the previously retrieved justifications. We detail these components in the next two sub-sections. 

\subsection{Iterative Justification Retrieval}
\label{retrieval_explanation}

\rs\ iteratively builds justification chains given a query. \rs\ starts by initializing the query with the concatenated question and candidate answer text\footnote{Note that this work can be trivially adapted to reading comprehension tasks. In such tasks (e.g., SQuAD~\cite{rajpurkar2018know}), the initial query would contain just the question text.}. Then, \rs\ iteratively repeats the following two steps: 
(a) It retrieves the most salient justification sentence given the current query using an alignment-IR approach\cite{yadavAHE}. The candidate justification sentences come from dataset-specific KBs. For example, in MultiRC, we use as candidates  all the sentences from the paragraph associated with the given question. In QASC, which has a large KB\footnote{In large KB-based QA, \rs\ first uses an off-the-shelf Lucene BM25\cite{BM25} to retrieve a pool of candidate justification sentences from which the evidence chains are constructed.} of 17.4 million sentences), similar to \citet{khot2019qasc} candidates are retrieved using the Heuristic+IR method which returns 80 candidate sentences for each candidate answer from the provided QASC KB.  
(b) it adjusts the query to focus on the {\em missing} information, i.e., the keywords that are not covered by the current evidence chain. \rs\ also dynamically adds new terms to the query from the previously retrieved justifications to nudge multi-hop retrieval. 
These two iterative steps repeat until a parameter-free termination condition is reached. 

We first detail the important components of \rs.

\begin{description}[topsep=0.2em, itemsep=0.1em, wide, labelwidth=!, labelindent=0pt]
\item[Alignment:] To compute the similarity score between a given query and a sentence from KB, \rs\ uses a vanilla unsupervised alignment method of \citet{yadavAHE} which uses only GloVe embeddings  \cite{pennington2014glove}.\footnote{Alignment based on BERT embeddings marginally outperformed the one based on GloVe embeddings, but BERT embeddings were much more expensive to generate.}
 The alignment method computes the cosine similarity between the word embeddings of each token in the query and each token in the given KB sentence, resulting in a matrix of cosine similarity scores. For each query token, the algorithm select the most similar token in the evidence text using max-pooling.
 At the end, the element-wise dot product between this max-pooled vector of cosine-similarity scores and the vector containing the IDF values of the query tokens is calculated to produce the overall alignment score $s$ for the given query $Q$ 
 and the supporting paragraph $P_j$: 
 \vspace{-2mm}
\begin{align}
s(Q, P_j) &= \sum_{i=1}^{|Q|} \mathit{idf}(q_i) \cdot \mathit{align}(q_i, P_j) \label{eqn:score}
\end{align}
\vspace{-5mm}
\begin{align}
\mathit{align}(q_i, P_j) &= \max_{k=1}^{|P_j|}\mathit{cosSim}(q_i, p_k) \label{eqn:align} 
\end{align}
where $q_i$ and $p_k$ are the $i^{th}$ and $k^{th}$ terms of the query ($Q$) and evidence sentence ($P_j$) respectively.

\item[Remainder terms ({\bf $Q_r$}):] Query reformulation in \rs\ is driven by the remainder terms, which are the set of query terms not yet covered in the justification set of $i$ sentences (retrieved from the first $i$ iterations of the retrieval process): 
\begin{align}
    Q_r(i) &= \displaystyle t(Q) - \bigcup_{s_k \in S_i} t(s_k)
    \label{Query_remainder}
\end{align}
where $t(Q)$ represents the unique set of query terms, $t(s_k)$ represents the unique terms of the $k^{th}$ justification, and $S_i$ represents the set of $i$ justification sentences. 
Note that we use soft matching of alignment for the inclusion operation: we consider a query term to be included in the set of terms in the justifications if its cosine similarity with a justification term is larger than a similarity threshold $M$ (we use $M$=0.95 for all our experiments - see \cref{adhoc_parameters}), thus ensuring that the two terms are similar in the embedding space.

\item[Coverage ($Q_c$):] measures the coverage of the query keywords by the retrieved chain of justifications $S$: 
\begin{align}
    Q_c(i) &= \displaystyle\frac{|\bigcup_{s_k \in S_i} t(Q) \cap t(s_k)|}{|t(Q)|}
    \label{Query_coverage}
\end{align}
where $|t(Q)|$ denotes the size of unique query terms. 

\end{description}

\paragraph{The \rs\ retrieval process}
\begin{enumerate}[wide, labelwidth=!, labelindent=0pt]

\item[Query reformulation:]  In each iteration $j$, \rs\ reformulates the query $Q(j)$ to include only the terms not yet covered by the current justification chain, $Q_r(j - 1)$. See, for example, the second hop in \cref{fig:AIR_multiRC_example}. 
To mitigate ambiguous queries, the query is expanded with the terms from all the previously retrieved justification sentences only if the number of uncovered terms is less than $T$ (we used $T = 2$ for MultiRC and $T = 4$ for QASC (see \cref{adhoc_parameters}). See, for example, the third hop in \cref{fig:AIR_multiRC_example}, in which the query is expanded with the terms of all the previously retrieved justification sentences. Formally:
\begin{align}
\tiny
Q(j) =
\begin{cases}
     Q_r(j-1),& \text{if } |Q_r(j-1)|>T\\
     Q_r(j-1) +  (t(s_{j-1}) - t(Q)),              & \text{otherwise} \\
\end{cases}
    \label{Query_reform}
\end{align}
where $j$ is the current iteration index.

\item[Stopping criteria:] \rs\ stops its iterative evidence retrieval process when either of the following conditions is true: 
(a) no new query terms are discovered in the last justification retrieved, i.e., $Q_r(i-1) == Q_r(i)$, or
(b) all query terms are covered by justifications, i.e., $Q_c =1$.
\end{enumerate}

\subsection{Answer Classification}

\rs's justification chains can be fed into any supervised answer classification method. For all experiments in this paper, we used RoBERTa~\cite{liu2019roberta}, a state-of-the-art transformer-based method. 
In particular, for MultiRC, we concatenate the query (composed from question and candidate answer text) with the evidence text, with the {\tt [SEP]} token between the two texts. A sigmoid is used over the {\tt [CLS]} representation to train a binary classification task\footnote{We used RoBERTa base with maximum sequence length of 512, batch size = 8, learning rate of 1e-5, and 5 number of epochs. RoBERTa-base always returned consistent performance on MultiRC experiments; many runs from RoBERTa-large failed to train (as explained by \cite{wolf2019transformers}), and generated near random performance.} (correct answer or not).


For QASC, we fine-tune RoBERTa as a multiple-choice QA \footnote{We used similar hyperparameters as in the MultiRC experiments, but instead used RoBERTa-large, with maximum sequence length of 128.} (MCQA)~\cite{wolf2019transformers} classifier with 8 choices using a softmax layer(similar to \cite{khot2019qasc}) instead of the sigmoid. The input text consists of eight queries (from eight candidate answers) and their corresponding eight evidence texts. Unlike the case of MultiRC, it is possible to train a MCQA classifier for QASC because every question has only 1 correct answer. We had also tried the binary classification approach for QASC but it resulted in nearly 5\% lower performance for majority of the experiments in \cref{tab:QASC}.

In QA tasks that rely on large KBs there may exist multiple chains of evidence that support a correct answer. This is particularly relevant in QASC, whose KB contains 17.2M facts.\footnote{The dataset creators make a similar observation~\cite{khot2019qasc}.} Figure~\ref{QASC_diagram} shows an example of this situation. To utilize this type of redundancy in answer classification, we extend \rs\ to extract {\em parallel evidence chains}. That is, to extract $N$ parallel chains, we run \rs\ $N$ times, ensuring that the first justification sentences in each chain are different (in practice, we start a new chain for each justification in the top $N$ retrieved sentences in the first hop). After retrieving $N$ parallel evidence chains, we take the union of all the individual justification sentences to create the supporting evidence text for that candidate answer.



\begin{table*}
\centering
\footnotesize
\setlength{\tabcolsep}{0.25em}
\begin{tabular}{@{} lllllllccc @{}}
\hline
\# & Computational  & Supervised & Method &  F1$_m$ & F1$_a$ & EM0 & \multicolumn{3} {c}{Evidence selection}   \\ 
  & steps & selection of & & & & & P & R & F1 \\
  & & justifications? \\
 \hline
 & & & \bf{DEVELOPMENT DATASET} \\
 \hline 
 & & & Baselines \\
 \hline 
 1 & $N$ & No & IR(paragraphs) \cite{khashabi2018looking} & 64.3 & 60.0 & 1.4 &  -- \\
 2 & $N$ & No & SurfaceLR \cite{khashabi2018looking} & 66.5 & 63.2 & 11.8 &  -- \\
 3 & $N$ & No & Entailment baseline \cite{trivedi2019repurposing} & 51.3 & 50.4 & -- & -- \\
\hline
& & & Previous work \\
 \hline 
4 & $N$ & Yes & QA+NLI \cite{pujari2019using} & - & - & 21.6 &  -- \\ 
5 & $N$ & Yes & EER$_{DPL}$ + FT \cite{wang2019evidence} & 70.5 & 67.8 & 13.3 &  -- \\
6 & $N$ & Yes & Multee (GloVe) \cite{trivedi2019repurposing} & 71.3 & 68.3 & 17.9 & -- \\
7 & $N$ & Yes & Multee (ELMo)$^\star$ \cite{trivedi2019repurposing} & 73.0 & 69.6 & 22.8 &  -- \\
8 & ${K\times N}$ & Yes & RS$^\star$ \cite{sun2018improving} & 73.1 & 70.5 & 21.8 & -- & -- & 60.8\\
9 & $N$ & No & BERT + BM25 \cite{yadav2019quick} & 71.1 & 67.4 & 23.1 & 43.8 & 61.2 & 51.0 \\
10 & $2^{N}-N-1$ & No & BERT + AutoROCC \cite{yadav2019quick} & 72.9 & 69.6 & 24.7 & 48.2 & 68.2 & 56.4 \\
\hline
& & & Alignment + RoBERTa(QA) baselines \\
	 \hline 
	11 & - & No & Entire passage + RoBERTa & 73.9 & 71.7 & 28.7 &17.4 & 100.0 & 29.6 \\
	12 & $N$ & No & Alignment ($k = 2$ sentences) + RoBERTa & 72.6 & 69.6  & 25.9 &  
62.4 & 55.6 & 58.8 \\
	13 & $N$ & No & Alignment ($k = 3$ sentences) + RoBERTa & 72.4 & 69.8  & 25.1 &  49.3 & 65.1 & 56.1 \\
	14 & $N$ & No & Alignment ($k = 4$ sentences) + RoBERTa & 73.6 & 71.4 & 28.0 &  41.0 & 72.0 & 52.3 \\
	15 & $N$ & No & Alignment ($k = 5$ sentences) + RoBERTa & 73.7 & 70.8 & 25.0 &  35.2 & 77.1 & 48.4 \\
\hline
& & & RoBERTa retriever + RoBERTa(QA) baselines \\
\hline
	16 & $N$ & Yes & RoBERTa-retriever(All passages) + RoBERTa & 70.5 & 68.0 & 24.9 & 63.4 & 61.1 & 62.3 \\
	17 & $N$ & Yes & RoBERTa-retriever(Fiction) + RoBERTa & 72.8 & 70.4 & 24.7 & 47.8 & 73.9 & 58.1 \\
	18 & $N$ & Yes & RoBERTa-retriever(News) + RoBERTa & 69.0 & 67.3 & 24.2 & 60.8 & 59.2 & 59.9\\
	19 & $N$ & Yes & RoBERTa-retriever(Science-textbook) + RoBERTa & 70.3 & 67.7 & 25.3 & 48.1 & 62.0 & 54.2 \\
	20 & $N$ & Yes & RoBERTa-retriever(Society\_Law) + RoBERTa & 72.8 & 70.3 & 25.3 & 50.4 & 68.5 & 58.0\\
	21 & ${K\times N}$ & Yes & RoBERTa-{\bf iterative}-retriever + RoBERTa & 70.1 & 67.6 & 24.0 & 67.1 & 58.4 & 62.5 \\
\hline
& & & RoBERTa + \rs\ (Parallel) Justifications & & &  \\
	 \hline 
	 22 & ${K\times N}$ & No & \rs\ (lexical) top chain + RoBERTa & 71.0 & 68.2 & 22.9 & 58.2 & 49.5 & 53.5\\
	23 &  ${K\times N}$ & No & \rs\ top chain + RoBERTa & \textit{\textbf{74.7}} & \textit{\textbf{72.3}} & \textit{\textbf{29.3}} & 66.2 & 63.1 & {\bf64.2} \\
	24 &  ${2 \times K\times N}$ & No & \rs\ Parallel evidence chains ($p = 2$) + RoBERTa & 75.5 & 73.6 & 32.5 & 50.4 & 71.9 & 59.2 \\
	25 & ${3 \times K\times N}$ & No & \rs\ Parallel evidence chains ($p = 3$) + RoBERTa & 75.8 & 73.7  & 30.6 & 40.8 & 76.7 & 53.3 \\
	26 & ${4 \times K\times N}$ & No & \rs\ Parallel evidence chains ($p = 4$) + RoBERTa & 76.3 & 74.2 & 31.3 & 34.8 & 80.8 & 48.7 \\
	27 & ${5 \times K\times N}$ & No & \rs\ Parallel evidence chains ($p = 5$) + RoBERTa & {\bf77.2} & {\bf75.1} & {\bf33.0} & 28.6 & 84.1 & 44.9 \\
\hline

\hline
& & & Ceiling systems with gold justifications \\
	 \hline
29 & - & Yes & EER$_{gt}$ + FT \cite{wang2019evidence} & 72.3 & 70.1 & 19.2 & --\\
30 & - & Yes & RoBERTa + Gold knowledge & 81.4 & 80 & 39 &  100.0 & 100.0 & 100.0\\
31 & - & - & Human & 86.4 & 83.8 & 56.6 &  -- \\
\hline
& & & \bf{TEST DATASET} \\
\hline
32 & $N$ & No & SurfaceLR \cite{khashabi2018looking} & 66.9 & 63.5 & 12.8\\
33 & $N$ & Yes & Multee (ELMo)$^\star$ \cite{trivedi2019repurposing} & 73.8 & 70.4 & 24.5 & -- \\
34 & $2^{N}-N-1$ & No & BERT + AutoROCC \cite{yadav2019quick} & {73.8} & {70.6} & {26.1}  \\
35 & ${5 \times K\times N}$ & No & RoBERTa + \rs\ (Parallel evidence = 5) & {\bf79.0} & {\bf76.4} & {\bf36.3} &   \\
\end{tabular}
\vspace{-0.5\baselineskip}
\caption{\small Results on the MultiRC development and test sets. 
	The first column specifies the runtime overhead required for selection of evidence sentences, where $N$ is the total number of sentences in the passage, and $K$ is the selected number of sentences. The second column specifies if the retrieval system is a supervised method or not. 
	The last three columns indicate evidence selection performance, whereas the previous three indicate overall QA performance.
	Only the last block of results report performance on the test set. 
	The bold italic font highlights the best performance without using parallel evidences. $\star$ denotes usage of external labeled data for pretraining.
	}

\label{tab:MultiRC}
\vspace{-5mm}
\end{table*}


\begin{table*}[t]
\centering
\footnotesize
\setlength{\tabcolsep}{0.25em}
\begin{tabular}{@{} lllrccl @{}}
\hline
\# & Number  & Method &  Accuracy  & Recall@10 & Recall@10 \\ 
	 & of steps &  & & \textit{both} & \textit{atleast one}  \\
	 & used? & & & \textit{found} & \textit{found} \\
\hline
 & & Baselines \\
\hline

0	& Single & 	Naive Lucene BM25 	& 35.6  & 17.2 & 68.1  \\
1	& Two &	Naive Lucene BM25 	& 36.3 & 27.8 & 65.7  \\
2	& Two & Heuristics+IR \cite{khot2019qasc}	& 32.4 & 41.6 & 64.4  \\
3	& - &	ESIM Q2Choice \cite{khot2019qasc} & 21.1 & 41.6 & 64.4  \\


\hline
& &  Previous work \\
\hline

4 & Single &  BERT-LC  \cite{khot2019qasc} & 59.8  & 11.7 & 54.7 \\ 

5 & Two &  BERT-LC  \cite{khot2019qasc} & 71.0 & 41.6 & 64.4 \\
6 & Two &  BERT-LC[WM]$^{\star}$ \cite{khot2019qasc} & 78.0 & 41.6 & 64.4 \\
\hline
  & & Alignment + RoBERTa baselines & & &  \\
\hline
7 & -- & No Justifiction + RoBERTa & 20.5 & 0 & 0  \\
	8 & Single & Alignment ($K = 1$ sentences) + RoBERTa & 54.4 & - & -  \\
	9 & Two &  Alignment ($K = 2$ sentences) + RoBERTa & 71.5  & - & -  \\
	10 & Two &  Alignment ($K = 3$ sentences) + RoBERTa & 73.3 & - & -  \\
	11 & Two &  Alignment ($K = 4$ sentences) + RoBERTa & 73.5  & - & -  \\
	12 & Two & Alignment ($K = 5$ sentences) + RoBERTa & 74.1 & - & -  \\
\hline
& & \rs + RoBERTa & & &  \\
\hline
13 & Two & \rs\ (lexical) top chain + RoBERTa & 75.8  & - & - \\
14 & Two & \rs\ top chain + RoBERTa & \textit{\textbf{76.2}}  & - & - \\
15 & Two & \rs\ Parallel evidence chains ($p = 2$) + RoBERTa & 79.8 & - & - \\
	16 & Two &  \rs\ Parallel evidence chains ($p = 3$) + RoBERTa & 80.9  & - & -  \\
	17 & Two & \rs\ Parallel evidence chains ($p = 4$) + RoBERTa & 79.7 &  - & - \\
	18 & Two &  \rs\ Parallel evidence chains ($p = 5$) + RoBERTa  & {\bf81.4} & {\bf44.8} & {\bf68.6}\\
\hline
& & \bf{TEST DATASET} \\
	 \hline
19 & Two &  BERT-LC \cite{khot2019qasc} & 68.5 & -  & - \\
20 & Two &  BERT-LC[WM]$^{\star}$ \cite{khot2019qasc} & 73.2 & - & - \\
21 & Two &  \rs\ Parallel evidence chains ($p = 5$) + RoBERTa & {\bf81.4} & - & -  \\
\end{tabular}
\vspace{-0.5\baselineskip}
\caption{\small QA and evidence selection performance on QASC. We also report recall@10 similar to \citet{khot2019qasc}. \textit{both found} reports the recall scores when both the gold justifications are found in top 10 ranked sentences and similarly \textit{atleast one found} reports the recall scores when either one or both the gold justifications are found in the top 10 ranked sentences. Recall@10 are not reported (row 8-17) when number of retrieved sentences are lesser than 10. Other notations are same as \cref{tab:MultiRC}. 
} 
\label{tab:QASC}
\end{table*}

\section{Experiments}

We evaluated our approach on two datasets:

{\flushleft {\bf Multi-sentence reading comprehension (MultiRC)}}, which is a reading comprehension dataset provided in the form of multiple-choice QA task~\cite{khashabi2018looking}. Every question is based on a paragraph, which contains the gold justification sentences for each question. We use every sentence of the paragraph as candidate justifications for a given question. Here we use the original MultiRC dataset,\footnote{\url{https://cogcomp.seas.upenn.edu/multirc/}} which includes the gold annotations for evidence text, unlike the version available on SuperGlue \cite{wang2019superglue}.  

{\flushleft {\bf Question Answering using Sentence Composition (QASC)}}, a large KB-based multiple-choice QA dataset~\cite{khot2019qasc}. Each question is provided with 8 answer candidates, out of which 4 candidates are hard adversarial choices. Every question is annotated with a fixed set of two justification sentences for answering the question. The justification sentences are to be retrieved from a KB having 17.2 million facts. As shown in the example of \cref{QASC_diagram} and also highlighted by \cite{khot2019qasc}, multiple evidence text are possible for a given question in QASC where the annotated gold justification sentences explain it more precisely. 

We report overall question answering performance as well as evidence selection performance in \cref{tab:MultiRC} for MultiRC, and \cref{tab:QASC} for QASC\footnote{\url{https://leaderboard.allenai.org/qasc/submissions/public}}.

\vspace{-1mm}
\subsection{Baselines}
In addition to previously-reported results, we include in the tables several in-house baselines.
For MultiRC, we considered three baselines. The first baseline is where we feed {\em all} passage sentences to the RoBERTa classifier (row 11 in \cref{tab:MultiRC}). The second baseline uses the alignment method of \cite{bridging_the_gap} to retrieve the top $k$ sentences ($k={2,5}$). 
Since \rs\ uses the same alignment approach for retrieving justifications in each iteration, the comparison to this second baseline highlights the gains from our iterative process with query reformulation.
The third baseline uses a supervised RoBERTa classifier trained to select the gold justifications  for every query (rows 16--21 in \cref{tab:MultiRC}). 
Lastly, we also developed a RoBERTa-based iterative retriever by concatenating the query with the retrieved justification in the previous step. We retrain the RoBERTa iterative retriever in every step, using the new query in each step. 

We considered two baselines for QASC. The first baseline does not include any justifications (row 7 in \cref{tab:QASC}). 
The second baseline uses the top $k$ sentences retrieved by the alignment method (row (8--12 in \cref{tab:QASC}).

\vspace{-1mm}

\subsection{Evidence Selection Results}

For evidence selection, we report precision, recall, and F1 scores on MultiRC (similar to \cite{wang2019evidence,yadav2019quick}). For QASC, we report Recall@10, similar to the dataset authors \cite{khot2019qasc}. 
We draw several observation from the evidence selection results:
\begin{enumerate}[label={\bf(\arabic*)}, topsep=0.2em, itemsep=0.1em, wide, labelwidth=!, labelindent=0pt]

\item {\bf \rs\ vs. unsupervised methods} - \rs\ outperforms all the unsupervised baselines and previous works in both MultiRC (row 9-15 vs. row 23 in table 1) and QASC(rows 0-6 vs. row 18). Thus, highlighting strengths of \rs\ over the standard IR baselines. \rs\ achieves 5.4\% better F1 score compared to the best parametric alignment baseline (row 12 in \cref{tab:MultiRC}), which highlights the importance of the  iterative approach over the vanilla alignment in \rs. Similarly, rows (4 and 5) of \cref{tab:QASC} also highlight this importance in QASC. 

\item {\bf \rs\ vs. supervised methods} - Surprisingly, \rs\ also outperforms the supervised RoBERTa-retriver in every setting(rows 16--21 in \cref{tab:MultiRC}). Note that the performance of this supervised retrieval method  drops considerably when trained on passages from a specific domain (row 19 in \cref{tab:MultiRC}), which highlights the domain sensitivity of supervised retrieval methods. In contrast, \rs\ is unsupervised and generalize better as it is not tuned to any specific domain. \rs\ also achieves better performance than supervised RoBERTa-iterative-retriever (row 21 in \cref{tab:MultiRC}) which simply concatenates the retrieved justification to the query after every iteration and further trains to retrieve the next justification. The RoBERTa-iterative-retriever achieves similar performance as that of the simple RoBERTa-retriever (row 16 vs. 21) which suggests that supervised iterative retrievers marginally exploit the information from query expansion. On the other hand, controlled query reformulation of \rs\ leads to 5.4\% improvement as explained in the previous point. 
All in all, \rs\ achieves state-of-the-art results for evidence retrieval on both MultiRC (row 23 in \cref{tab:MultiRC}) and QASC (row 18 of \cref{tab:QASC}).

\item {\bf Soft-matching of \rs} - the alignment-based \rs\ is 10.7\% F1 better than \rs\ that relies on lexical matching (rather than the soft matching) on MultiRC (row 22 vs. 23), which emphasizes the advantage of alignment methods over conventional lexical match approaches.

\end{enumerate}

\subsection{Question Answering Results}

For overall QA performance, we report the standard performance measures ($F1_a$, $F1_m$, and $EM0$) in MultiRC \cite{khashabi2018looking}, and accuracy for QASC~\cite{khot2019qasc}. 

The results in tables~\ref{tab:MultiRC} and~\ref{tab:QASC} highlight:

\begin{enumerate}[label={\bf(\arabic*)}, topsep=0.2em, itemsep=0.1em, wide, labelwidth=!, labelindent=0pt]

\item {\bf State-of-the-art performance:} \\
{\bf Development set} - On both MultiRC and QASC, RoBERTa fine-tuned using the \rs\ retrieved evidence chains (row 23 in \cref{tab:MultiRC} and row 14 in \cref{tab:QASC}) outperforms all the previous approaches and the baseline methods. This indicates that the evidence texts retrieved by \rs\  not only provide better explanations, but also contribute considerably in achieving the best QA performance.

{\bf Test set} - On the official hidden test set, RoBERTa fine-tuned on 5 parallel evidences from \rs\ achieves new state-of-the-art QA results, outperforming previous state-of-the-art methods by 7.8\% accuracy on QASC (row 21 vs. 20), and 10.2\% EM0 on MultiRC (row 35 vs. 34).

\item {\bf Knowledge aggregation} - The knowledge aggregation from multiple justification sentences leads to substantial improvements, particularly in QASC (single justification (row 4 and 8) vs. evidence chains (row 5 and row 9) in table \cref{tab:QASC}). Overall, the chain of evidence text retrieved by \rs\ enables knowledge aggregation resulting in the improvement of QA performances. 

\item {\bf Gains from parallel evidences} - Further, knowledge aggregation from parallel evidence chains lead to another 3.7\% EM0 improvement on MultiRC (row 27), and 5.6\% on QASC over the single \rs\ evidence chain (row 18). To our knowledge, these are new state-of-the-art results in both the datasets. 

\end{enumerate}

\begin{table}
    \small
    \setlength{\tabcolsep}{0.25em}
    \begin{tabular}{@{} c|c|c|c|c @{}}
   \# of & BM25 & \rs  & Alignment &\rs  \\
 hops  & & (Lexical) & & \textit{uncontrolled}  \\
   & & \textit{uncontrolled} & &   \\
    \hline
 1 & 38.8 & 38.8 & 46.5 & 46.5 \\
 2 & 48.4 &  45.9 & 58.8  & 54.1  \\ 
 3 & 48.4 & 45.8 & 56.1 & 52.2  \\
 4 & 47.0 & 44.0 & 52.3 & 49.1  \\
 5 & 44.8 & 41.1 & 48.4 & 46.0  \\

    \end{tabular}
     \caption{\small{Impact of semantic drift across consecutive hops on justification selection F1-performance of MultiRC development set. The \textit{uncontrolled} configuration indicates that the justification sentences retrieved in each hop were appended to the query in each step. Here, \rs\ is forced to retrieve the same  number of justifications as indicated by the \# of hops.}}
    \label{tab:Semantic_drift}
    \end{table}

\vspace{-3mm}

\section{Analysis}

To further understand the retrieval process of \rs\ we implemented several analyses. 

\subsection{Semantic Drift Analysis}
\label{semantic_drift_analysis}

To understand the importance of modeling missing information in query reformulation, we analyzed a simple variant of \rs\ in which, rather the focusing on missing information, we simply concatenate the complete justification sentence to the query after each hop. 
To expose semantic drift, we retrieve a specified number of justification sentences. 
As seen in \cref{tab:Semantic_drift}, now the \rs(lexical)-\textit{uncontrolled} and \rs-\textit{uncontrolled} perform worse than both BM25 and the alignment method. This highlights that the focus on missing information during query reformulation is an important deterrent of semantic drift. 
We repeated the same experiment with the supervised RoBERTa retriever (trained iteratively for 2 steps) and the original parameter-free \rs, which decides its number of hops using the stopping conditions. Again, we observe similar performance drops in both: the  RoBERTa retriever drops from 62.3\% to 57.6\% and \rs\ drops to 55.4\%.

\begin{table}
    \centering
    \small
    \setlength{\tabcolsep}{0.25em}
    \begin{tabular}{@{} c|c|cc @{}}
   $Q_r$ & MultiRC & \multicolumn{2} {c}{QASC}  \\
   & F1 score & Both Found & One Found \\
    \hline
 1 & {\bf 64.2} & 41.7 & 67.7 \\
 2 & 62.7 & 42.7 & 67.7 \\
 3 & 61.8 & 43.1 & 68.6 \\
 4 & 60.63 & 40.6 & 68.4 \\
 5 & 59.8 & 39.0 & 67.5
    \end{tabular}
    \caption{\small{Impact on justification selection F1-performance from the hyper parameter $Q_r$ of \rs\ (\cref{Query_reform}).}}
    \label{tab:qr}
    \end{table}

\subsection{Robustness to Hyper Parameters}
\label{adhoc_parameters}
We evaluate the sensitivity of \rs\ to the 2 hyper parameters: the threshold ($Q_r$) for query expansion, and the cosine similarity threshold $M$ in computation of alignment. As shown in \cref{tab:adhoc_param_sensitivity}, 
evidence selection performance of \rs\ drops with the lower values of $M$ but the drops are small, suggesting that \rs\ is robust to different $M$ values. 

Similarly, there is a drop in performance for MultiRC with the increase in the $Q_r$ threshold used for query expansion, hinting to the occurrence of semantic drift for higher values of $Q_r$ (\cref{tab:qr}). 
This is because the candidate justifications are coming from a relatively small numbers of paragraphs in MultiRC; thus even shorter queries ($=2$ words) can retrieve relevant justifications. On the other hand, the number of candidate justifications in QASC is much higher, which requires longer queries for disambiguation ($>= 4$ words).



    \begin{table}
     \begin{tabular}{@{} c|c|cc @{}}
   $M$ & MultiRC & \multicolumn{2} {c}{QASC}  \\
   & F1 score & Both Found & One Found \\
    \hline
 0.95 & {\bf 64.2} & 43.1 & 68.6 \\
 0.85 & 63.7 & 42.3 & 67.9 \\
 0.75 & 63.4 & 42.5 & 68.0 \\
    \end{tabular}
    \caption{\small{Impact on justification selection F1 score from the hyper parameter $M$ in the alignment step (\cref{retrieval_explanation}).}}
    \label{tab:adhoc_param_sensitivity}
    \vspace{-5mm}
\end{table}

\subsection{Saturation of Supervised Learning}

To verify if the MultiRC training data is sufficient to train a supervised justification retrieval method, we trained justification selection classifiers based on BERT, XLNet, and RoBERTa on increasing proportions of the MultiRC training data (\cref{tab:ablations}). 
This analysis indicates that all three classifiers approach their best performance at around 5\% of the training data. This indicates that, while these supervised methods converge quickly, they are unlikely to outperform \rs, an unsupervised method, even if more training data were available.

\section{Conclusion}
We introduced a simple, unsupervised approach for evidence retrieval for question answering. Our approach combines three ideas: (a) an unsupervised alignment approach to soft-align questions and answers with justification sentences using GloVe embeddings, (b) an iterative process that reformulates queries focusing on terms that are not covered by existing justifications, and (c) a simple stopping condition that concludes the iterative process when all terms in the given question and candidate answers are covered by the retrieved justifications. Overall, despite its simplicity, unsupervised nature, and its sole reliance on GloVe embeddings, our approach outperforms all previous methods (including supervised ones) on the evidence selection task on two datasets: MultiRC and QASC. When these evidence sentences are fed into a RoBERTa answer classification component, we achieve the best QA performance on these two datasets. Further, we show that considerable improvements can be obtained by aggregating knowledge from parallel evidence chains retrieved by our method.

In addition of improving QA, we hypothesize that these simple unsupervised components of \rs\ will benefit future work on supervised neural iterative retrieval approaches by improving their query reformulation algorithms and termination criteria.

\begin{table}
    \centering
    \small
    \setlength{\tabcolsep}{0.25em}
    \begin{tabular}{@{} c|c|c|c|c @{}}
  \%\ of training data & BERT  & XLnet & RoBERTa & \rs\   \\
    \hline
  2 & 55.2  & 54.6 & 62.3 &   \\
  5 & 60.0  & 59.6 & 60.8 &   \\ 
  10 & 59.9 & 57.0 & 59.8 &   \\
  15 & 58.3 & 59.9  & 59.1 &   \\
  20 & 58.5 & 60.2  & 60.0 & 64.2  \\
  40 & 58.5 & 58.7 & 58.8 &   \\
  60 & 59.1 & 61.4 & 59.8 &   \\
  80 & 59.3 & 61.0 & 60.5 &   \\
  100 & 60.9 & 61.1 & 62.3 &   \\
    \end{tabular}
    \caption{\small Comparison of \rs\ with XLNet, RoBERTa, and BERT on justification selection task, trained on increasing proportion of the training data on MultiRC. }
    \label{tab:ablations}
    \vspace{-4mm}
\end{table}

\section*{Acknowledgments}
We thank Tushar Khot (AI2) and Daniel Khashabhi (AI2) for helping us with the dataset and evaluation resources. 
This work was supported by the Defense Advanced Research Projects Agency
(DARPA) under the World Modelers program, grant number
W911NF1810014.
Mihai Surdeanu declares a financial interest in lum.ai. This interest has been properly disclosed to the University of Arizona Institutional Review Committee and is managed in accordance with its conflict of interest policies.


\bibliographystyle{acl_natbib}
\bibliography{acl2020}

\begin{thebibliography}{48}
\expandafter\ifx\csname natexlab\endcsname\relax\def\natexlab#1{#1}\fi

\bibitem[{Alvarez-Melis and Jaakkola(2017)}]{alvarez2017causal}
David Alvarez-Melis and Tommi Jaakkola. 2017.
\newblock A causal framework for explaining the predictions of black-box
  sequence-to-sequence models.
\newblock In \emph{Proceedings of the 2017 Conference on Empirical Methods in
  Natural Language Processing}, pages 412--421.

\bibitem[{Arras et~al.(2017)Arras, Horn, Montavon, M{\"u}ller, and
  Samek}]{arras2017relevant}
Leila Arras, Franziska Horn, Gr{\'e}goire Montavon, Klaus-Robert M{\"u}ller,
  and Wojciech Samek. 2017.
\newblock " what is relevant in a text document?": An interpretable machine
  learning approach.
\newblock \emph{PloS one}, 12(8):e0181142.

\bibitem[{Banerjee(2019)}]{banerjee2019asu}
Pratyay Banerjee. 2019.
\newblock Asu at textgraphs 2019 shared task: Explanation regeneration using
  language models and iterative re-ranking.
\newblock In \emph{Proceedings of the Thirteenth Workshop on Graph-Based
  Methods for Natural Language Processing (TextGraphs-13)}, pages 78--84.

\bibitem[{Bauer et~al.(2018)Bauer, Wang, and Bansal}]{bauer2018commonsense}
Lisa Bauer, Yicheng Wang, and Mohit Bansal. 2018.
\newblock Commonsense for generative multi-hop question answering tasks.
\newblock In \emph{Proceedings of the 2018 Conference on Empirical Methods in
  Natural Language Processing}, pages 4220--4230.

\bibitem[{Biran and Cotton(2017)}]{biran2017explanation}
Or~Biran and Courtenay Cotton. 2017.
\newblock Explanation and justification in machine learning: A survey.
\newblock In \emph{IJCAI-17 workshop on explainable AI (XAI)}, volume~8.

\bibitem[{Chen et~al.(2017)Chen, Fisch, Weston, and Bordes}]{chen2017reading}
Danqi Chen, Adam Fisch, Jason Weston, and Antoine Bordes. 2017.
\newblock Reading wikipedia to answer open-domain questions.
\newblock In \emph{Proceedings of the 55th Annual Meeting of the Association
  for Computational Linguistics (Volume 1: Long Papers)}, pages 1870--1879.

\bibitem[{Chen and Durrett(2019)}]{chen2019understanding}
Jifan Chen and Greg Durrett. 2019.
\newblock Understanding dataset design choices for multi-hop reasoning.
\newblock In \emph{Proceedings of the 2019 Conference of the North American
  Chapter of the Association for Computational Linguistics: Human Language
  Technologies, Volume 1 (Long and Short Papers)}, pages 4026--4032.

\bibitem[{Chen et~al.(2019)Chen, Lin, and Durrett}]{chen2019multi}
Jifan Chen, Shih-ting Lin, and Greg Durrett. 2019.
\newblock Multi-hop question answering via reasoning chains.
\newblock \emph{arXiv preprint arXiv:1910.02610}.

\bibitem[{Choi et~al.(2017)Choi, Hewlett, Uszkoreit, Polosukhin, Lacoste, and
  Berant}]{choi2017coarse}
Eunsol Choi, Daniel Hewlett, Jakob Uszkoreit, Illia Polosukhin, Alexandre
  Lacoste, and Jonathan Berant. 2017.
\newblock Coarse-to-fine question answering for long documents.
\newblock In \emph{Proceedings of the 55th Annual Meeting of the Association
  for Computational Linguistics (Volume 1: Long Papers)}, pages 209--220.

\bibitem[{Das et~al.(2018)Das, Dhuliawala, Zaheer, and McCallum}]{das2018multi}
Rajarshi Das, Shehzaad Dhuliawala, Manzil Zaheer, and Andrew McCallum. 2018.
\newblock Multi-step retriever-reader interaction for scalable open-domain
  question answering.

\bibitem[{Devlin et~al.(2019)Devlin, Chang, Lee, and
  Toutanova}]{devlin2018bert}
Jacob Devlin, Ming-Wei Chang, Kenton Lee, and Kristina Toutanova. 2019.
\newblock Bert: Pre-training of deep bidirectional transformers for language
  understanding.
\newblock In \emph{Proceedings of the 2019 Conference of the North American
  Chapter of the Association for Computational Linguistics: Human Language
  Technologies, Volume 1 (Long and Short Papers)}, pages 4171--4186.

\bibitem[{Dua et~al.(2019)Dua, Wang, Dasigi, Stanovsky, Singh, and
  Gardner}]{dua2019drop}
Dheeru Dua, Yizhong Wang, Pradeep Dasigi, Gabriel Stanovsky, Sameer Singh, and
  Matt Gardner. 2019.
\newblock Drop: A reading comprehension benchmark requiring discrete reasoning
  over paragraphs.
\newblock In \emph{Proceedings of the 2019 Conference of the North American
  Chapter of the Association for Computational Linguistics: Human Language
  Technologies, Volume 1 (Long and Short Papers)}, pages 2368--2378.

\bibitem[{Feldman and El-Yaniv(2019)}]{feldman2019multi}
Yair Feldman and Ran El-Yaniv. 2019.
\newblock Multi-hop paragraph retrieval for open-domain question answering.
\newblock In \emph{Proceedings of the 57th Annual Meeting of the Association
  for Computational Linguistics}, pages 2296--2309.

\bibitem[{Geva and Berant(2018)}]{geva2018learning}
Mor Geva and Jonathan Berant. 2018.
\newblock Learning to search in long documents using document structure.
\newblock In \emph{Proceedings of the 27th International Conference on
  Computational Linguistics}, pages 161--176.

\bibitem[{Gilpin et~al.(2018)Gilpin, Bau, Yuan, Bajwa, Specter, and
  Kagal}]{gilpin2018explaining}
Leilani~H Gilpin, David Bau, Ben~Z Yuan, Ayesha Bajwa, Michael Specter, and
  Lalana Kagal. 2018.
\newblock Explaining explanations: An overview of interpretability of machine
  learning.
\newblock In \emph{2018 IEEE 5th International Conference on Data Science and
  Advanced Analytics (DSAA)}, pages 80--89. IEEE.

\bibitem[{Gravina et~al.(2018)Gravina, Rossetto, Severini, and
  Attardi}]{gravina2018cross}
Alessio Gravina, Federico Rossetto, Silvia Severini, and Giuseppe Attardi.
  2018.
\newblock Cross attention for selection-based question answering.
\newblock In \emph{2nd Workshop on Natural Language for Artificial
  Intelligence}. Aachen: R. Piskac.

\bibitem[{Jansen and Ustalov(2019)}]{jansen2019textgraphs}
Peter Jansen and Dmitry Ustalov. 2019.
\newblock Textgraphs 2019 shared task on multi-hop inference for explanation
  regeneration.
\newblock In \emph{Proceedings of the Thirteenth Workshop on Graph-Based
  Methods for Natural Language Processing (TextGraphs-13)}, pages 63--77.

\bibitem[{Khashabi et~al.(2018{\natexlab{a}})Khashabi, Chaturvedi, Roth,
  Upadhyay, and Roth}]{khashabi2018looking}
Daniel Khashabi, Snigdha Chaturvedi, Michael Roth, Shyam Upadhyay, and Dan
  Roth. 2018{\natexlab{a}}.
\newblock Looking beyond the surface: A challenge set for reading comprehension
  over multiple sentences.
\newblock In \emph{Proceedings of the 2018 Conference of the North American
  Chapter of the Association for Computational Linguistics: Human Language
  Technologies, Volume 1 (Long Papers)}, pages 252--262.

\bibitem[{Khashabi et~al.(2018{\natexlab{b}})Khashabi, Khot, Sabharwal, and
  Roth}]{khashabi2018question}
Daniel Khashabi, Tushar Khot, Ashish Sabharwal, and Dan Roth.
  2018{\natexlab{b}}.
\newblock Question answering as global reasoning over semantic abstractions.
\newblock In \emph{Thirty-Second AAAI Conference on Artificial Intelligence}.

\bibitem[{Khot et~al.(2019{\natexlab{a}})Khot, Clark, Guerquin, Jansen, and
  Sabharwal}]{khot2019qasc}
Tushar Khot, Peter Clark, Michal Guerquin, Peter Jansen, and Ashish Sabharwal.
  2019{\natexlab{a}}.
\newblock Qasc: A dataset for question answering via sentence composition.
\newblock \emph{arXiv preprint arXiv:1910.11473}.

\bibitem[{Khot et~al.(2019{\natexlab{b}})Khot, Sabharwal, and
  Clark}]{khot2019missing}
Tushar Khot, Ashish Sabharwal, and Peter Clark. 2019{\natexlab{b}}.
\newblock What’s missing: A knowledge gap guided approach for multi-hop
  question answering.
\newblock In \emph{Proceedings of the 2019 Conference on Empirical Methods in
  Natural Language Processing and the 9th International Joint Conference on
  Natural Language Processing (EMNLP-IJCNLP)}, pages 2807--2821.

\bibitem[{Kim et~al.(2017)Kim, Fiorini, Wilbur, and Lu}]{bridging_the_gap}
Sun Kim, Nicolas Fiorini, W~John Wilbur, and Zhiyong Lu. 2017.
\newblock Bridging the gap: Incorporating a semantic similarity measure for
  effectively mapping pubmed queries to documents.
\newblock \emph{Journal of biomedical informatics}, 75:122--127.

\bibitem[{Lin et~al.(2018)Lin, Ji, Liu, and Sun}]{lin2018denoising}
Yankai Lin, Haozhe Ji, Zhiyuan Liu, and Maosong Sun. 2018.
\newblock Denoising distantly supervised open-domain question answering.
\newblock In \emph{Proceedings of the 56th Annual Meeting of the Association
  for Computational Linguistics (Volume 1: Long Papers)}, pages 1736--1745.

\bibitem[{Liu et~al.(2019)Liu, Ott, Goyal, Du, Joshi, Chen, Levy, Lewis,
  Zettlemoyer, and Stoyanov}]{liu2019roberta}
Yinhan Liu, Myle Ott, Naman Goyal, Jingfei Du, Mandar Joshi, Danqi Chen, Omer
  Levy, Mike Lewis, Luke Zettlemoyer, and Veselin Stoyanov. 2019.
\newblock Roberta: A robustly optimized bert pretraining approach.
\newblock \emph{arXiv preprint arXiv:1907.11692}.

\bibitem[{Min et~al.(2018)Min, Zhong, Socher, and Xiong}]{min2018efficient}
Sewon Min, Victor Zhong, Richard Socher, and Caiming Xiong. 2018.
\newblock Efficient and robust question answering from minimal context over
  documents.
\newblock \emph{arXiv preprint arXiv:1805.08092}.

\bibitem[{Nie et~al.(2019)Nie, Wang, and Bansal}]{nie2019revealing}
Yixin Nie, Songhe Wang, and Mohit Bansal. 2019.
\newblock Revealing the importance of semantic retrieval for machine reading at
  scale.
\newblock In \emph{Proceedings of the 2019 Conference on Empirical Methods in
  Natural Language Processing and the 9th International Joint Conference on
  Natural Language Processing (EMNLP-IJCNLP)}, pages 2553--2566.

\bibitem[{Pennington et~al.(2014)Pennington, Socher, and
  Manning}]{pennington2014glove}
Jeffrey Pennington, Richard Socher, and Christopher Manning. 2014.
\newblock Glove: Global vectors for word representation.
\newblock In \emph{Proceedings of the 2014 conference on empirical methods in
  natural language processing (EMNLP)}, pages 1532--1543.

\bibitem[{Pujari and Goldwasser(2019)}]{pujari2019using}
Rajkumar Pujari and Dan Goldwasser. 2019.
\newblock Using natural language relations between answer choices for machine
  comprehension.
\newblock In \emph{Proceedings of the 2019 Conference of the North American
  Chapter of the Association for Computational Linguistics: Human Language
  Technologies, Volume 1 (Long and Short Papers)}, pages 4010--4015.

\bibitem[{Qi et~al.(2019)Qi, Lin, Mehr, Wang, and Manning}]{qi2019answering}
Peng Qi, Xiaowen Lin, Leo Mehr, Zijian Wang, and Christopher~D Manning. 2019.
\newblock Answering complex open-domain questions through iterative query
  generation.
\newblock In \emph{Proceedings of the 2019 Conference on Empirical Methods in
  Natural Language Processing and the 9th International Joint Conference on
  Natural Language Processing (EMNLP-IJCNLP)}, pages 2590--2602.

\bibitem[{Rajpurkar et~al.(2018)Rajpurkar, Jia, and Liang}]{rajpurkar2018know}
Pranav Rajpurkar, Robin Jia, and Percy Liang. 2018.
\newblock Know what you don’t know: Unanswerable questions for squad.
\newblock In \emph{Proceedings of the 56th Annual Meeting of the Association
  for Computational Linguistics (Volume 2: Short Papers)}, pages 784--789.

\bibitem[{Robertson et~al.(2009)Robertson, Zaragoza et~al.}]{BM25}
Stephen Robertson, Hugo Zaragoza, et~al. 2009.
\newblock The probabilistic relevance framework: Bm25 and beyond.
\newblock \emph{Foundations and Trends{\textregistered} in Information
  Retrieval}, 3(4):333--389.

\bibitem[{Speer et~al.(2017)Speer, Chin, and Havasi}]{speer2017conceptnet}
Robyn Speer, Joshua Chin, and Catherine Havasi. 2017.
\newblock Conceptnet 5.5: An open multilingual graph of general knowledge.
\newblock In \emph{AAAI}, pages 4444--4451.

\bibitem[{Sun et~al.(2019{\natexlab{a}})Sun, Bedrax-Weiss, and
  Cohen}]{sun2019pullnet}
Haitian Sun, Tania Bedrax-Weiss, and William Cohen. 2019{\natexlab{a}}.
\newblock Pullnet: Open domain question answering with iterative retrieval on
  knowledge bases and text.
\newblock In \emph{Proceedings of the 2019 Conference on Empirical Methods in
  Natural Language Processing and the 9th International Joint Conference on
  Natural Language Processing (EMNLP-IJCNLP)}, pages 2380--2390.

\bibitem[{Sun et~al.(2019{\natexlab{b}})Sun, Yu, Chen, Yu, Choi, and
  Cardie}]{sun2019dream}
Kai Sun, Dian Yu, Jianshu Chen, Dong Yu, Yejin Choi, and Claire Cardie.
  2019{\natexlab{b}}.
\newblock Dream: A challenge data set and models for dialogue-based reading
  comprehension.
\newblock \emph{Transactions of the Association for Computational Linguistics},
  7:217--231.

\bibitem[{Sun et~al.(2019{\natexlab{c}})Sun, Yu, Yu, and
  Cardie}]{sun2018improving}
Kai Sun, Dian Yu, Dong Yu, and Claire Cardie. 2019{\natexlab{c}}.
\newblock Improving machine reading comprehension with general reading
  strategies.
\newblock In \emph{Proceedings of the 2019 Conference of the North American
  Chapter of the Association for Computational Linguistics: Human Language
  Technologies, Volume 1 (Long and Short Papers)}, pages 2633--2643.

\bibitem[{Surdeanu et~al.(2008)Surdeanu, Ciaramita, and
  Zaragoza}]{surdeanu2008learning}
Mihai Surdeanu, Massimiliano Ciaramita, and Hugo Zaragoza. 2008.
\newblock Learning to rank answers on large online qa collections.
\newblock In \emph{Proceedings of ACL-08: HLT}, pages 719--727.

\bibitem[{Sydorova et~al.(2019)Sydorova, Poerner, and
  Roth}]{sydorova2019interpretable}
Alona Sydorova, Nina Poerner, and Benjamin Roth. 2019.
\newblock Interpretable question answering on knowledge bases and text.
\newblock \emph{arXiv preprint arXiv:1906.10924}.

\bibitem[{Trivedi et~al.(2019)Trivedi, Kwon, Khot, Sabharwal, and
  Balasubramanian}]{trivedi2019repurposing}
Harsh Trivedi, Heeyoung Kwon, Tushar Khot, Ashish Sabharwal, and Niranjan
  Balasubramanian. 2019.
\newblock Repurposing entailment for multi-hop question answering tasks.
\newblock In \emph{Proceedings of the 2019 Conference of the North American
  Chapter of the Association for Computational Linguistics: Human Language
  Technologies, Volume 1 (Long and Short Papers)}, pages 2948--2958.

\bibitem[{Tu et~al.(2019)Tu, Huang, Wang, Huang, He, and Zhou}]{tu2019select}
Ming Tu, Kevin Huang, Guangtao Wang, Jing Huang, Xiaodong He, and Bowen Zhou.
  2019.
\newblock Select, answer and explain: Interpretable multi-hop reading
  comprehension over multiple documents.
\newblock \emph{arXiv preprint arXiv:1911.00484}.

\bibitem[{Wang et~al.(2019{\natexlab{a}})Wang, Pruksachatkun, Nangia, Singh,
  Michael, Hill, Levy, and Bowman}]{wang2019superglue}
Alex Wang, Yada Pruksachatkun, Nikita Nangia, Amanpreet Singh, Julian Michael,
  Felix Hill, Omer Levy, and Samuel~R Bowman. 2019{\natexlab{a}}.
\newblock Superglue: A stickier benchmark for general-purpose language
  understanding systems.
\newblock \emph{arXiv preprint arXiv:1905.00537}.

\bibitem[{Wang et~al.(2019{\natexlab{b}})Wang, Yu, Sun, Chen, Yu, McAllester,
  and Roth}]{wang2019evidence}
Hai Wang, Dian Yu, Kai Sun, Jianshu Chen, Dong Yu, David McAllester, and Dan
  Roth. 2019{\natexlab{b}}.
\newblock Evidence sentence extraction for machine reading comprehension.
\newblock In \emph{Proceedings of the 23rd Conference on Computational Natural
  Language Learning (CoNLL)}, pages 696--707.

\bibitem[{Welbl et~al.(2018)Welbl, Stenetorp, and
  Riedel}]{welbl2018constructing}
Johannes Welbl, Pontus Stenetorp, and Sebastian Riedel. 2018.
\newblock Constructing datasets for multi-hop reading comprehension across
  documents.
\newblock \emph{Transactions of the Association of Computational Linguistics},
  6:287--302.

\bibitem[{Wolf et~al.(2019)Wolf, Debut, Sanh, Chaumond, Delangue, Moi, Cistac,
  Rault, Louf, Funtowicz et~al.}]{wolf2019transformers}
Thomas Wolf, Lysandre Debut, Victor Sanh, Julien Chaumond, Clement Delangue,
  Anthony Moi, Pierric Cistac, Tim Rault, R{\'e}mi Louf, Morgan Funtowicz,
  et~al. 2019.
\newblock Transformers: State-of-the-art natural language processing.
\newblock \emph{arXiv preprint arXiv:1910.03771}.

\bibitem[{Yadav et~al.(2019{\natexlab{a}})Yadav, Bethard, and
  Surdeanu}]{yadavAHE}
Vikas Yadav, Steven Bethard, and Mihai Surdeanu. 2019{\natexlab{a}}.
\newblock Alignment over heterogeneous embeddings for question answering.
\newblock In \emph{Proceedings of the 2019 Conference of the North American
  Chapter of the Association for Computational Linguistics: Human Language
  Technologies, (Long Papers)}, Minneapolis, USA. Association for Computational
  Linguistics.

\bibitem[{Yadav et~al.(2019{\natexlab{b}})Yadav, Bethard, and
  Surdeanu}]{yadav2019quick}
Vikas Yadav, Steven Bethard, and Mihai Surdeanu. 2019{\natexlab{b}}.
\newblock Quick and (not so) dirty: Unsupervised selection of justification
  sentences for multi-hop question answering.
\newblock In \emph{Proceedings of the 2019 Conference on Empirical Methods in
  Natural Language Processing and the 9th International Joint Conference on
  Natural Language Processing (EMNLP-IJCNLP)}, pages 2578--2589.

\bibitem[{Yadav et~al.(2018)Yadav, Sharp, and Surdeanu}]{yadav2018sanity}
Vikas Yadav, Rebecca Sharp, and Mihai Surdeanu. 2018.
\newblock Sanity check: A strong alignment and information retrieval baseline
  for question answering.
\newblock In \emph{The 41st International ACM SIGIR Conference on Research \&
  Development in Information Retrieval}, pages 1217--1220. ACM.

\bibitem[{Yang et~al.(2019)Yang, Dai, Yang, Carbonell, Salakhutdinov, and
  Le}]{yang2019xlnet}
Zhilin Yang, Zihang Dai, Yiming Yang, Jaime Carbonell, Russ~R Salakhutdinov,
  and Quoc~V Le. 2019.
\newblock Xlnet: Generalized autoregressive pretraining for language
  understanding.
\newblock In \emph{Advances in neural information processing systems}, pages
  5754--5764.

\bibitem[{Yang et~al.(2018)Yang, Qi, Zhang, Bengio, Cohen, Salakhutdinov, and
  Manning}]{yang2018hotpotqa}
Zhilin Yang, Peng Qi, Saizheng Zhang, Yoshua Bengio, William Cohen, Ruslan
  Salakhutdinov, and Christopher~D Manning. 2018.
\newblock Hotpotqa: A dataset for diverse, explainable multi-hop question
  answering.
\newblock In \emph{Proceedings of the 2018 Conference on Empirical Methods in
  Natural Language Processing}, pages 2369--2380.

\end{thebibliography}
\end{document}